\documentclass[journal,  onecolumn]{IEEEtran}

\def\ps@headings{%
\def\@oddhead{\mbox{}\scriptsize\rightmark \hfil \thepage}%
\def\@evenhead{\scriptsize\thepage \hfil \leftmark\mbox{}}%
\def\@oddfoot{}%
\def\@evenfoot{}}

\IEEEoverridecommandlockouts

\usepackage{cite,url}
\usepackage{amsmath}
\usepackage{amssymb}
\usepackage{mathrsfs}
\usepackage{nameref}
\usepackage[inline]{enumitem}
\usepackage{soul}

\usepackage[dvipsnames]{xcolor}
\usepackage{latexsym}
\usepackage{verbatim} 

\usepackage[center]{caption}
\usepackage{algorithmic}
\usepackage{algorithm}

\ifCLASSINFOpdf
\usepackage[pdftex]{graphicx}
  \graphicspath{Figs/}
\DeclareGraphicsExtensions{.pdf,.jpeg,.png}
\else
\usepackage[dvips]{graphicx}
\fi

\usepackage{epstopdf}      
\usepackage{todonotes}
\usepackage{color}
\usepackage{hyperref}
\hypersetup{
    colorlinks=false,
    filecolor=magenta,      
    urlcolor=blue,
}

\author{
\IEEEauthorblockN{
Denis Kleyko\IEEEauthorrefmark{1},
Ross W. Gayler\IEEEauthorrefmark{2}, 
Evgeny Osipov\IEEEauthorrefmark{3}\\ 
}
\IEEEauthorblockA{
\IEEEauthorrefmark{1}Research Institutes of Sweden, Kista, Sweden and University of California, Berkeley, USA.
denis.kleyko@ri.se}\\
\IEEEauthorblockA{\IEEEauthorrefmark{2}Independent researcher, Melbourne, Australia. ross@rossgayler.com} \\
\IEEEauthorblockA{\IEEEauthorrefmark{3}Lule\aa{} University of Technology, Lule\aa{}, Sweden. evgeny.osipov@ltu.se} \\
}
\begin{document}

\title{
{
Commentaries on ``Learning Sensorimotor Control with Neuromorphic Sensors: Toward Hyperdimensional Active Perception'' [Science Robotics Vol. 4 Issue 30 (2019) 1--10]
}
}




\maketitle

\begin{abstract}

This correspondence comments on the findings reported in a recent Science Robotics article by Mitrokhin et al.~\cite{Mitrokhin19HD}. 
The main goal of this commentary is to expand on some of the issues touched on in that article.
Our experience is that hyperdimensional computing is very different from other approaches to computation and that it can take considerable exposure to its concepts before attaining practically useful understanding.
Therefore, in order to provide an overview of the area to the first time reader of~\cite{Mitrokhin19HD}, the commentary includes a brief historic overview as well as connects the findings of the article to a larger body of literature existing in the area.

\end{abstract}

\section{Introduction}

The recent article by A. Mitrokhin, P. Sutor, C. Ferm\"{u}ller, and Y. Aloimonos, 
``Learning Sensorimotor Control with Neuromorphic Sensors: Toward Hyperdimensional Active Perception'', 
which appeared in Science Robotics vol. 4 issue 30 (2019), presents a case for using a computation framework called hyperdimensional computing also known as Vector Symbolic Architectures (VSAs) for fusing motoric abilities of a robot with its perception system. 
The idea of computing with random vectors as basic objects is also known as 
Holographic Reduced Representation~\cite{PlateTr}, 
Multiply-Add-Permute~\cite{MAP}, 
Binary Spatter Codes~\cite{Kanerva97},
Binary Sparse Distributed Codes~\cite{CDT2001},
Matrix Binding of Additive Terms~\cite{Gallant13}, and
Semantic Pointer Architecture~\cite{BuildBrain}.
All these frameworks are essentially equivalent. 
In the light of the present very high level of attention to the area of autonomous AI-empowered systems from the industry and the society, we hope and believe that the application of VSAs in robotics will get an appropriately increasing attention from the community of AI/robotics researchers and practitioners. 
Our own experience with VSAs has shown that due to its considerable difference from the conventional computing paradigms the development of intuition and understanding required for practical applications needs to be supported by extended exposure to the details and interpretation of VSAs. 
We, therefore, write this commentary with the aim to facilitate the readers of the original article to comprehend and exploit the very different perspective on computation that is both required and enabled by VSAs. 
Hopefully this will inspire more exciting applications of VSAs to robotics.

The commentary is organized as follows. Section~\ref{sect:digest} presents a compact summary of the main contributions in the original article.
Section~\ref{sect:VSAs} provides a brief historical excursus into VSAs as well as their current state with respect to applications areas.
Peculiarities of representing data in VSAs are discussed in Section~\ref{sec:VSAs:data:repr}.
Section~\ref{sec:VSAs:capacity} concludes the commentary by discussing the information capacity of VSAs.

\section{A digest of the original article}
\label{sect:digest}
The main focus of the original article is in addressing the problem of active perception, where an agent's knowledge 
emerges as the result of an interplay between the agent's actions and the sensory input arising/causing those actions.
It is proposed to use VSA's computational framework to jointly represent the sensory data and the agent's motor/action information taken to generate these data.
The authors of the article convey a message that  VSAs  
are beneficial to the area of robotics as means for implementing active perception. The article delivers two major technical observations:
\begin{itemize}
\item 
VSA-based representation of different sensor modalities enables formation and flexible manipulation of memory sequences (time-series) of the sensory data where parts of the representation could be easily modified by either inserting or deleting some sensory data;

\item A joint representation of sensory data and motor/action information using VSAs enables a more streamlined interface to conventional machine learning architectures and results in faster and more resource-efficient learning with comparable accuracy.
\end{itemize}

VSAs operate with high-dimensional vectors (HD vectors); typically, with several thousands of elements. They are also referred to in the original article as HBV representations or distributed representations. 
The foundation for the solution described in the article is a variant of VSAs based on dense binary HD vectors, which was originally introduced by P. Kanerva~\cite{Kanerva09}.
In this variant, elements of HD vectors take only binary values ``0'' or ``1''. 
The computations in VSAs are based on compositions of three simple arithmetic operations on HD vectors: \textit{binding} - implemented as elementwise exclusive OR (XOR) and denoted as $\oplus$; \textit{bundling} - implemented as consensus sum/majority rule; and \textit{permutation} of the elements.
VSAs also require a similarity metric. The similarity between two binary HD vectors is characterized by the normalized Hamming distance (denoted as $\Delta_h$). It measures the rate of the number of element positions in the two compared HD vectors ($\textbf{a}$ and $\textbf{b}$) in which they have different element values: 
\begin{equation}
\label{eq:hamming}
\Delta_h(\textbf{a},\textbf{b})= \frac{\lVert  \textbf{a} \oplus \textbf{b}
\rVert_{1}}{N} = \frac{\sum^{N}_{i=1}\textbf{a}_i \oplus \textbf{b}_i}{N}.
\end{equation}

The authors of the original article identify these arithmetic operations with semantics specific to the  article's focus. 
The binding operation was used in two contexts: to construct the representation of an unordered set of items and to construct the representation of the assignment of a value to a variable.
The permutation operation was used to construct the representation of the position of an an item in an ordered sequence. 
Finally, the bundling operation was used to construct the representation of a composite data structure, which is essentially a set of assignments of values to variables. 
It is important to understand that this identification of VSAs operations with semantics is for expository convenience in the original article and is not an essential part of VSAs.
We present an extended discussion of some different usages of the same fundamental arithmetic operations in Section~\ref{sec:VSAs:data:repr}.

In essence, the solution proposed by the authors consists of the following steps:
\begin{enumerate}
  \item Representation of an entire visual frame as one HD vector, 
  using a chain of binding operations on HD vectors corresponding to pixel intensities. The pixel intensity vectors are permuted to associate the pixels' positions in the frame.
  \item A procedure for generating a set of HD vectors to represent the range of pixel intensities. The similarities between the generated HD vectors preserve the underlying similarities between the pixel intensity values. 
  This part is not highlighted by the authors as a specific contribution of the article, since it is based on the technique previously proposed by them.
  \item Representation of a visual sequence as a memory HD vector.
  The frames represented as HD vectors produced at step 1 are assigned to HD vectors representing the time ticks using the binding operation.
  This produces an HD vector that represents the assignment of a value (the HD vector of a visual frame)
  to a variable (the HD vector corresponding to a time tick).
  The whole visual sequence is represented in an HD vector storing a set of all the HD vectors representing value-variable pairs in the sequence.
  This HD vector is calculated as the bundling operation on all HD vectors for value-variable pairs.
  \item A pipeline for learning sensorimotor relationships, that is the correspondence of the sensory data (visual sequence) to the motor/action producing it. 
  This is achieved by producing a joint HD vector of a visual frame and the value of robot's velocity, under which the frame was observed. 
This joint representation was used in the  heteroassociative memory mode, which allowed obtaining a prediction of the target velocity for a given visual frame.
\end{enumerate}

In fact, it is even possible to interpret and implement VSAs operations in terms of standard artificial neural networks
(ANN).
However, this interpretation risks raising unhelpful expectations.
For example, unlike typical ANNs, an ANN implementing the VSAs operations (i.e., binding, bundling, and permutation) has fixed connection weights and consequently no need for a process to incrementally optimize those weights.
``Learning'' in a circuit composed from the VSAs operations is often implemented by bundling (effectively, a single step of addition),
but may also be implemented as updating an autoassociative memory or Sparse Distributed Memory~\cite{KanervaBook},
or even an update of a non-neural content-addressable memory.
Consequently, it is probably more helpful to think of the VSAs operations as signal processing transformations that work on very high-dimensional signals in order to manipulate discrete data structures represented by vectors.

As a transient result in the original article the authors briefly report on an experiment where a fully connected ANN effectively implemented the final step for decoding the distributed representations reflecting that one could get comparable accuracy while avoiding the complexity of training convolutional ANN. 
The authors further highlighted as their main finding that matching accuracy could be achieved with pure single-pass feed-forward operations by querying the heteroassociative memory storing HD vectors of  scenes bound with their velocities.
While this is definitely a valid approach we emphasize the importance of a scenario where conventional ANNs (e.g., fully connected ones) are trained by an iterative process using HD vectors as input's representation. 
Our own experiments with distributed representations of texts and using them as an input to conventional machine learning classifiers for solving natural language processing tasks~\cite{HyperEmbed} shown that it is possible to achieve substantial speedups of the training and operation phases as well as to get significant reduction of memory consumption. 

One of the contributions highlighted by the authors of the original article is the mapping of sensory data from a neuromorphic dynamic vision sensor camera into a VSAs representation. 
In essence, they represent the averaged event statistics as pixel intensities over a predefined time interval. 
The resulting frame is then treated as a one channel visual frame.

Finally, along the way to their technical solution, the authors present the theoretical
findings on the capacity of the memory in HD vectors. Specifically, the authors computed the expected normalized Hamming distance for an HD vector storing the result of the bundling operation (please see the details in Section~\ref{sec:VSAs:capacity}).

\section{Vector Symbolic Architectures: historical notes and the current state}
\label{sect:VSAs}


VSAs is an umbrella term for a family of computational frameworks 
using high-dimensional distributed representations~\cite{Hinton1986}
and relying on the algebraic properties provided by a set of operators (bundling, binding, and permutation)
on the high-dimensional representation space.
VSAs are intimately related to tensor product binding, introduced by P. Smolensky~\cite{Smolensky1990}.
His method uses the tensor product as the binding operator,
which results in the dimensionality of the resultant vector being the product of the dimensionalities of the operand vectors.
The key points demonstrated were that it is possible to represent complex composite data structures (usually thought of as symbolic data structures) in a vector space,
and that it is possible to define transformations on that vector space, which manipulate the represented composite structures without needing to decompose them.

VSA binding operations can be interpreted as forming the tensor product of the operands and then projecting that result back into a vector space with the same dimensionality as each of the operand vectors.
Consequently, the dimensionality of the representational space remains constant,
whereas the dimensionality of tensor product binding representations increases exponentially with the number of terms being bound.
As a family, the VSAs have been committed to using a fixed dimensionality vector space for the representation and manipulation of composite data structures
by exploiting the algebraic properties of a small number of operators on that vector space.
Specific VSAs frameworks were introduced by
T. Plate~\cite{PlateTr}, P. Kanerva~\cite{Kanerva97}, R. Gayler~\cite{MAP}, D. Rachkovskij~\cite{Rachkovskij2001}, and others~\cite{GAHRR, Gallant13, Laiho} in the time span of the 1990s to 2000s. 
The interested readers are kindly referred to~\cite{VSAcomp} where eight different VSAs frameworks are compared and taxonomized. 
The inspirations of these authors vary, but precursors of these works can be found in, at least, convolutional models of human memory~\cite{murdock_theory_1982,metcalfe_eich_composite_1982} 
and models of associative memories inspired by holography~\cite{willshaw_non-holographic_1969,borsellino_convolution_1973,schonemann_algebraic_1987}.

Starting off with a neurophysiological inspiration that cognitive functions in biological brains are achieved through highly parallel activations of a large number of neurons, 
which through the learning process form statistically persistent patterns, 
VSAs offer a computational model where everything (items/entities/concepts/symbols/scalars etc.) is represented by HD vectors, 
which act as a distributed representation.  
VSAs provide an extreme form of distributed representation in that it is holographic and stochastic.
The holographic property means that the information content of the representation is spread over all the dimensions of the high-dimensional space.
Any subset of the dimensions can be used as a noisy approximation to the full set of dimensions.
The representations have to be interpreted stochastically, in that the relative proximities of vectors (as measured by the angles between them) are the important properties
and the value of any specific dimension makes only a small contribution to the direction of a vector and is consequently nearly irrelevant (in isolation).
This is the major departure from the standpoints of the conventional computing, 
where each component of a representation has a different specific meaning
and the exact value of each component is generally important.



Compared to standard ANNs and more recent neural techniques such as Deep Learning the number of people working on VSAs, and, therefore, the number of publications, has been exceptionally low.
One possible explanation for this disparity is that VSAs are very different from other computational approaches,
so it can take a considerable investment of effort for a researcher to become sufficiently familiar to appreciate the potential advantages.
Compounding this, VSAs can be implemented by ANNs and have often been presented as a type of neural network,
but because VSAs are so different from standard ANNs
the prior knowledge brought from standard ANNs may actually hinder understanding VSAs.
Another, possibly cynical, explanation is that VSAs are hard relative to standard ANNs precisely because they do not rely on optimization of a large number of connection weights.
When a VSAs system works in some task it does so because of good design of the system, not by throwing vast amounts of data and computational resources at an optimization process.

However, the level of VSAs activity has been increasing over the years.
Partly, this is due to cumulation of interesting use cases.
Researchers have demonstrated the utility of VSAs in different applications (see them at the end of this section),
which then serves as encouragement and inspiration for future researchers.

In recent years the level of interest in VSAs has increased dramatically.
The main driver of this has been the realization that VSAs may be well suited to the next generation of computing hardware.
Digital computing hardware has been constantly shrinking in size, but is now reaching the scale where individual computational elements will be inherently unreliable.
Also, although hardware can be made massively parallel, the software capability to effectively use that parallelism is badly lagging.
Because of the holographic and stochastic properties of VSAs
it may be ideally suited as a basis for computation on massively parallel, unreliable hardware~\cite{HDNP17}.
Consequently there has been a dramatic upsurge of research on hardware implementation of VSAs~\cite{LiHD16, LiHD17, Schmuck2019}.


Due to their nature, VSAs could be interpreted as a framework allowing manipulations of discrete structures with analog computing. 
Indeed, as we will see in Section~\ref{sec:VSAs:data:repr}, mapping discrete structures to HD vectors is not a very complicated problem, though it has its caveats.
Moreover, the usage of high-dimensionality provides the representations with very high robustness against errors (e.g., bit flips) appearing during the computation. 
This has a counter-intuitive effect well-studied in stochastic computing~\cite{ComputingRandomness}, which is a related six decades old computational paradigm, that certain computational tasks can be implemented more efficiently compared to the conventional computing on exact digital devices. 
Indeed, the intuition fails easily at this claim. 
How come that computing with several hundred dimensional vectors could be more efficient then computing with $16$, $32$ or $64$ bits only?
A short answer here is that the robustness to noise allows working on approximate analog hardware, which intrinsically introduces noise in the results of computations.
This noise could be devastating for computations assuming precise values of individual positions, which is not the case in VSAs.
In particular, there are recent developments in hardware~\cite{MemristorHD19}, which suggest designs based on memristive technologies, tailored to VSAs operations using only a fraction of energy needed for the conventional computer architectures.
Another prominent recent work~\cite{TPAM} demonstrates how representations of a particular VSAs framework -- Frequency Domain Holographic Reduced Representations~\cite{PlateBook} -- can be mapped to spike timing codes, which opens possibilities for implementing VSAs on neuromorphic hardware.




In general, despite being relatively little-known outside of a small community of researchers, VSAs have a broad range of possible applications.  
Since the whole area was always close to cognitive science many of VSAs use-cases spans different sub-areas, of AI such as cognitive architectures~\cite{Eliasmith2012}, analogical reasoning~\cite{Eliasmith11, RachkovskijAnalogy, Emruli13, Levy14}, word embedding~\cite{RI_first, RPRSKJ2015, Widdows15} and $n$-gram statistics~\cite{RIJHK2015} for natural language processing.  
Also, since recently, there are numerous applications of VSAs for machine learning~\cite{Rasanen2014, Rasanen2015tr, ACCESS_BIOFAULT, HDGestureIEEE}. 
AI, however, is not the only application area for VSAs.
There are other areas of computer science where VSAs could be useful, for example, for implementing conventional data structures such as finite-state machines~\cite{HD_FSA, UCBHD_FSA}.
Randomized algorithms is another such area as it was recently shown that a data structure for approximate membership query known as Bloom Filter~\cite{TPBF} is a subclass of VSAs~\cite{ABF}. 
There are even examples of applying VSAs to such areas as communications~\cite{Jakimovski, KimHDM} and workflow scheduling~\cite{HDWorkflow}.


When it comes to the applications of VSAs in the area of robotics, an early work~\cite{AAAIWLevyBajracharyaGaylor} was to use VSAs to program robot's reactive behavior.
Nevertheless, even today, the number of applications of VSAs in robotics is fairly limited. 
At the same time, the combination of the two enabling factors above: energy efficient hardware and broad functionality makes VSAs very perspective for robotic applications.
We see the confirmation of this promise in the amount of the recently observed activity. 
For example, it is quite staggering that there is a very interesting recent work~\cite{Neubert2019}, which appeared almost at the same time as the article~\cite{Mitrokhin19HD}.
The work~\cite{Neubert2019} provides an introduction to VSAs for robotics as well as overviews the use of VSAs for different robotic tasks.
In particular, these tasks were:
\begin{itemize}
    \item recognition of objects from multiple viewpoints, which is important for mobile robot localization by recognizing known landmarks;
    \item  recognition of objects for manipulation, and other robotics tasks;
    \item  learning and recall of reactive behavior~\cite{VSAsRB}; 
    \item sequence processing for place recognition by implementing a SLAM algorithm with VSAs.

\end{itemize}

Thus, given the broad spectrum of possible applications, we completely agree with the authors that VSAs hold a promise for robotics, machine learning, AI, and computer science, in general.

\section{Data representation in VSAs}
\label{sec:VSAs:data:repr}

\subsection{Atomic representations}

When designing a VSAs-based system for solving a problem it is common to define a set of the most basic items/entities/ concepts/symbols/scalars for the given problem and assign them with HD vectors, which are referred to as atomic HD vectors.
The process of assigning atomic HD vectors is often referred to as mapping\footnote{Sometimes such terms as projection and embedding are also used.}.
In the early days of VSAs, most of the works were focusing on symbolic problems. 
In the case of working with symbols or characters one could easily imagine many tasks where a reasonable assumption would be that symbols are not related at all and, therefore, their atomic HD vectors can be generated at random. 
A recent example of using this assumption for creating a useful VSAs algorithm is a method  of mapping conventional $n$-gram statistics into an HD vector~\cite{RIJHK2015}.
On the other hand, it was also realized that there are many problems where assigning atomic HD vectors randomly does not lead to any useful behavior of the designed system.
Therefore, in general, for a given problem we need to choose atomic representations of items such that the similarity between the representations of items correspond to the properties that we care about and want to drive the system dynamics.

In practice, numerical values are data type, which is present in many problems, e.g., in the area of machine learning.
Arguably, representing numerical values with random HD vectors is not necessarily the best option, therefore, it is worth considering similarity preserving HD vectors.
As mentioned in~\cite{Mitrokhin19HD}, one way to do it is to directly embed the linear relationship between numerical values in HD vectors, which is, indeed, a known approach for initiating similarity preserving atomic HD vectors (see, e.g.,~\cite{Widdows15, HD_ICRC16} and Appendix I in~\cite{HDGestureIEEE}).
It is also worth mentioning that there has been proposed a number of other mapping techniques~\cite{Scalarencoding} for representing numerical values as HD vectors. 
These mappings (including the linear one) were shown to perform well on some classification problems~\cite{TNNLS18, HDGestureIEEE}.
Another work~\cite{Purdy16} presented numerous similarity preserving techniques (encoders) for mapping different data types to sparse binary HD vectors. 
These techniques can be even generalized to the case of dense binary HD vectors. 
There is also a promising mapping presented in~\cite{Rasanen2015con} for forming real-valued HD vectors from numerical vectors.

\begin{figure}[ht]
\begin{center}
\includegraphics[width=0.8\columnwidth]{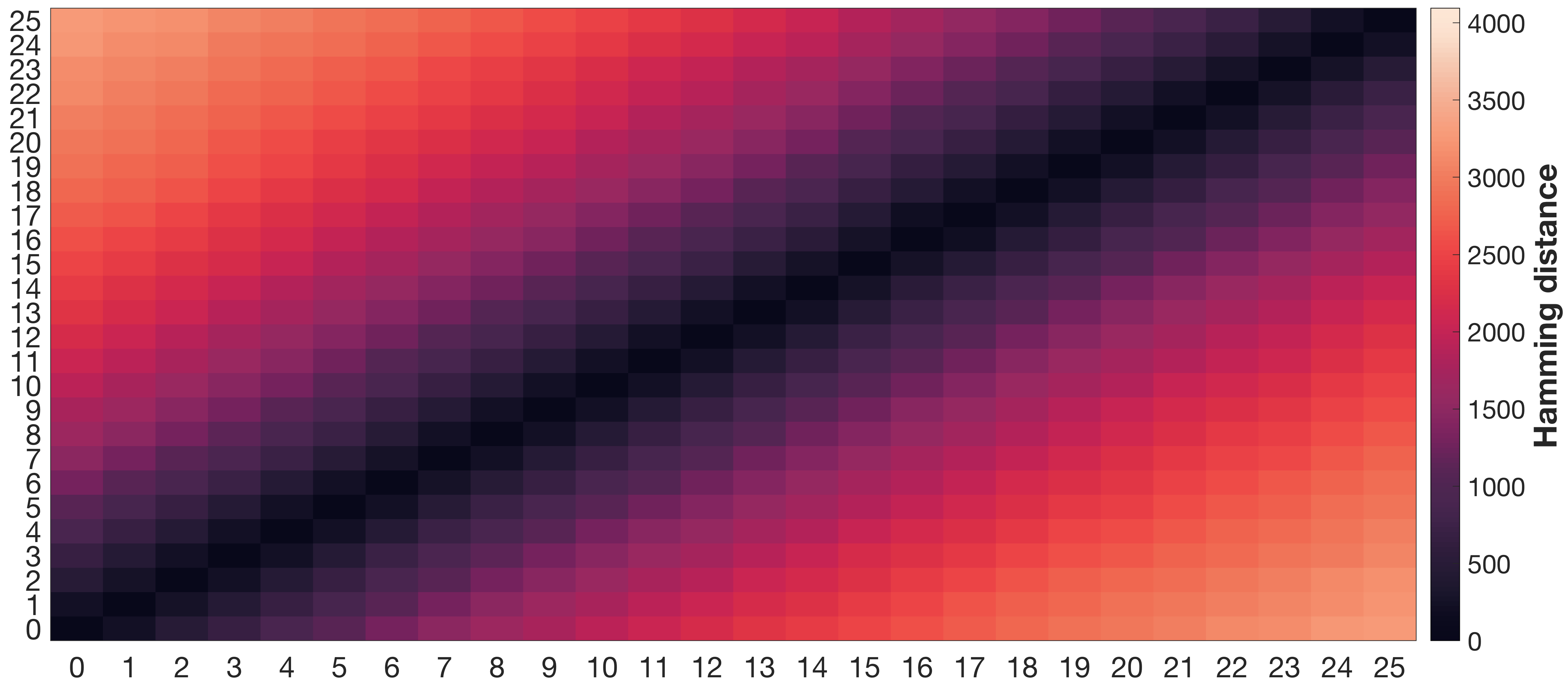}
\caption{
Hamming distance visualization of intensities. 
Visualization of the Hamming distance between intensity values in the range of 0 to 25. 
The heat map is obtained for $8192$-dimensional HD vectors using the nonlinear mapping~\cite{Scalarencoding}.
The proximity parameter of the mapping was set to $0.03$.
} 
\label{fig:heatmap} 
\end{center}
\end{figure}

On the other hand, if there is data associated with a problem an alternative way to mappings would be to obtain atomic HD vectors with the help of the available data via, e.g., an optimization process. 
For example, in~\cite{Mitrokhin19HD} the optimization-based mapping~\cite{Sutor18} was used. 
Fig. 3 in~\cite{Mitrokhin19HD} exemplifies the heat map of the normalized Hamming distance structure for the representations of intensities obtained using the optimization-based mapping.
It is interesting that a similar distance structure can be obtained using simpler methods such as a nonlinear mapping~\cite{Scalarencoding}. 
The corresponding heat map is presented in Fig.~\ref{fig:heatmap}.  
Similar to Fig. 3 in~\cite{Mitrokhin19HD}, distances away from the diagonal are increasing but the difference to the optimization-based mapping is that the nonlinear mapping is more ``aggressive'' in increasing the distances for the neighboring intensities.
This observation does not neglect the importance of optimization-based representations.
It rather shows that similar distance structures can be obtained in different ways.

Last, it is worth mentioning that the optimization-based method~\cite{Sutor18} for obtaining similarity preserving HD vectors can be contrasted to a Random Indexing method~\cite{RI_first}.
Random Indexing also implicitly (i.e., without constructing the co-occurrence matrix) uses co-occurrence statistics in order to form similarity preserving HD vectors based on available data. The difference, however, is that Random Indexing is optimization-free as it forms representations through a single pass via the data, and, thus, it can be done  on-line while the optimization required by~\cite{Sutor18} calls for off-line processing.
It is an interesting research question whether these methods arrive at similar representations in terms of inter-item distance structure or whether the optimization process brings extra benefits by refining the inter-item distance structure.


\subsection{Composite representations}



The atomic representations per se are hardly sufficient to solve a problem.
Instead, they are used as a building block in forming a composite representations.
Besides atomic HD vectors, when constructing composite representations a set of operations for manipulating HD vectors is also used.  
During this process, as a designer you are free to choose between different ways of both forming atomic HD vectors and combining them via the known operations. 
The main guiding principle, however, is that your choice of constructing composite representations should result in such similarity structure that would support the properties necessary for the problem at hand. 
Nevertheless, neither choice is right nor wrong -- this illustrates that the choice of representation only has to be compatible with how it will be used, since it is like having a choice of data structures in conventional programming.

As emphasized in~\cite{Neubert2019}, in VSAs there is no structured (w.r.t. to the methodology) way for designing systems.
It is also true that the area is missing well-defined design patterns.
On the other hand, one could argue that there are best practices, which are commonly used when building a solution and could be seen as a set of available design choices.   
For example, for the task of testing membership in a set, there is a well-known data structure called Bloom filter, which has been shown to be a special case of VSAs~\cite{ABF} where the bundling operation is implemented via OR and atomic HD vectors are sparse. 
In~\cite{Mitrokhin19HD}, an alternative way of representing a set through a chain of binding operations involving all atomic HD vectors was used. 
When using such design choice it is important to keep in mind that if there are similar HD vectors representing different members of the set these HD vectors will largely cancel each other when  XOR is used for binding.
Representation of ordered pairs or ordered sequences via the use of binding and permutation operations is another notable example of VSAs best practices~\cite{Kanerva09}.
For instance, it has proven to be useful for representing sequential information when forming VSAs-based word embeddings~\cite{RPRSKJ2015}.
As mentioned in the article, an interesting property of an HD vector representing the ordered sequence is that a single permutation applied to this vector can shift the whole sequence either forward or backward.
We also believe that this interesting phenomenon is a potentially powerful design choice.
An example of using this phenomenon is an application of shifting the HD vector for the ordered sequence  by permutation for searching the best alignment of two sequences~\cite{BidTrans14}.
The article also proposed a new design choice for forming a composite representation of a scene model for dynamic vision sensors.
Dynamic vision sensors typically send a stream of visual ``events'' corresponding to a series of fixations rather than representing a scene as a raster image.
However, until today, it is not very clear how to represent such streams in the context of VSAs. 
Therefore, transforming the stream to an image, which is then represented as an HD vector, allows mitigating this issue since now a scene is a ``bag of structured events''.
The proposed design choice is different from existing choices~\cite{HRRimage, Gallant2016, WeissOlshausen16} for images, as~\cite{HRRimage} focuses on storing images in HD vectors, while~\cite{Gallant2016, WeissOlshausen16, FracScene} proposed choices which preserve local similarity for nearby locations in an image. 
In contrast to~\cite{Gallant2016, WeissOlshausen16, FracScene}, the proposed design choice of forming  HD vectors of images would produce very different representations even in the case when intensities of only a single pixel are very different (i.e., their corresponding HD vectors are quasi-orthogonal). 
One could imagine that such choice would not be translation invariant, 
which is a very important property  in computer vision applications. 
Translation invariant mapping of images is a challenging problem. 
Only few works indirectly and partially address this problem for VSAs. 
The work~\cite{Neubert2019} proposes to map all possible rotations of an object with a given step of rotation degree in a single composite HD vector representing an object. 
Similar ideas could also be found in the Map Seeking Circuit approach~\cite{MSC}. 
The ideas used in the Map Seeking Circuits might possibly be implemented with VSAs using the mappings as in\cite{WeissOlshausen16, FracScene} to form a composite HD vector for, e.g., several copies of the same object at different spatial locations. 
Although, as mentioned above the mapping technique presented in the original article is just a possible design choice - one should be aware of the possible limitations in computer vision tasks. 

Moreover, a list of the design choices is not exhaustive and new mechanisms are being discovered. 
For example, the task of probing a component from an HD vector representing an ordered pair is relatively simple task.
This task becomes much more complex (complexity grows exponentially) in the case of either ordered sequence or when a set is represented via a chain of binding operations.
However, there is a very recent work~\cite{Resonator}, which proposed an elegant mechanism (i.e., a design choice) called Resonator Circuit to address this problem. 
Other interesting examples of newly discovered design choices are works~\cite{VecDer, FracBind} where new ways of using and implementing the binding operation were proposed.
In~\cite{FracBind}, a new fractional binding was used to represent continuous spaces while in~\cite{VecDer} a new binding for Holographic Reduced Representations~\cite{PlateTr} was presented.
To conclude, it should be admitted that there is unlikely one correct way of designing a VSAs-based solution to a problem.
Therefore, when solving any problem with VSAs the resulting solution is a choice, not the choice.

\subsection{Composite representations for active perception}

One of the motivations for the article is the so-called ``active perception'' and the fact that actions and perceptions are often stored separately. 
An intuitive way of achieving ``active perception'' would be to focus on the perception, which is ``conditional'' on the action taken. 
In the context of VSAs, this could be translated to representing the sensory  information (i.e., perception) or its change as the binding with a representation of the action, thus, forming composite HD vector, which acts as a sensorimotor representation.
For instance, when considering VSAs representation of frames of role-filler pairs~\cite{Kanerva97}, roles can represent arbitrarily complex structures\footnote{Unlike classic AI where roles are atomic symbols.} (e.g., perceptions)  while fillers can represent actions, hence,  the whole representation could be seen as a sensorimotor program.
Thus, it is fair to say that the notion of sensorimotor representations in the from of HD vectors has been highlighted in the VSAs literature from the beginning. 
For example, Appendix~I in~\cite{PlateThesis}  deals with the representation of numbers in the context of arithmetic tables where numbers and operands can be seen as perception while the result of the operation as an associated action. 
As additional results supporting the idea of forming sensorimotor representation via binding HD vectors for sensory and motor/action information, it is worth mentioning works~\cite{AAAIWLevyBajracharyaGaylor, Bees, EmruliBICA}.
In~\cite{AAAIWLevyBajracharyaGaylor}, VSAs were used to program robot's behavior.
In~\cite{Bees}, VSAs were used to form scene representation for experiments with honey bees. 
It was shown that VSAs can be used in order to mimic the process of learning concepts using the process of reinforcement.   
Work~\cite{EmruliBICA} proposed a VSA-based approach for learning behaviors based on observing and associating sensory and motor/action information.


\section{Information capacity of VSAs}
\label{sec:VSAs:capacity}

\begin{figure}[ht]
\begin{center}
\includegraphics[width=0.8\columnwidth]{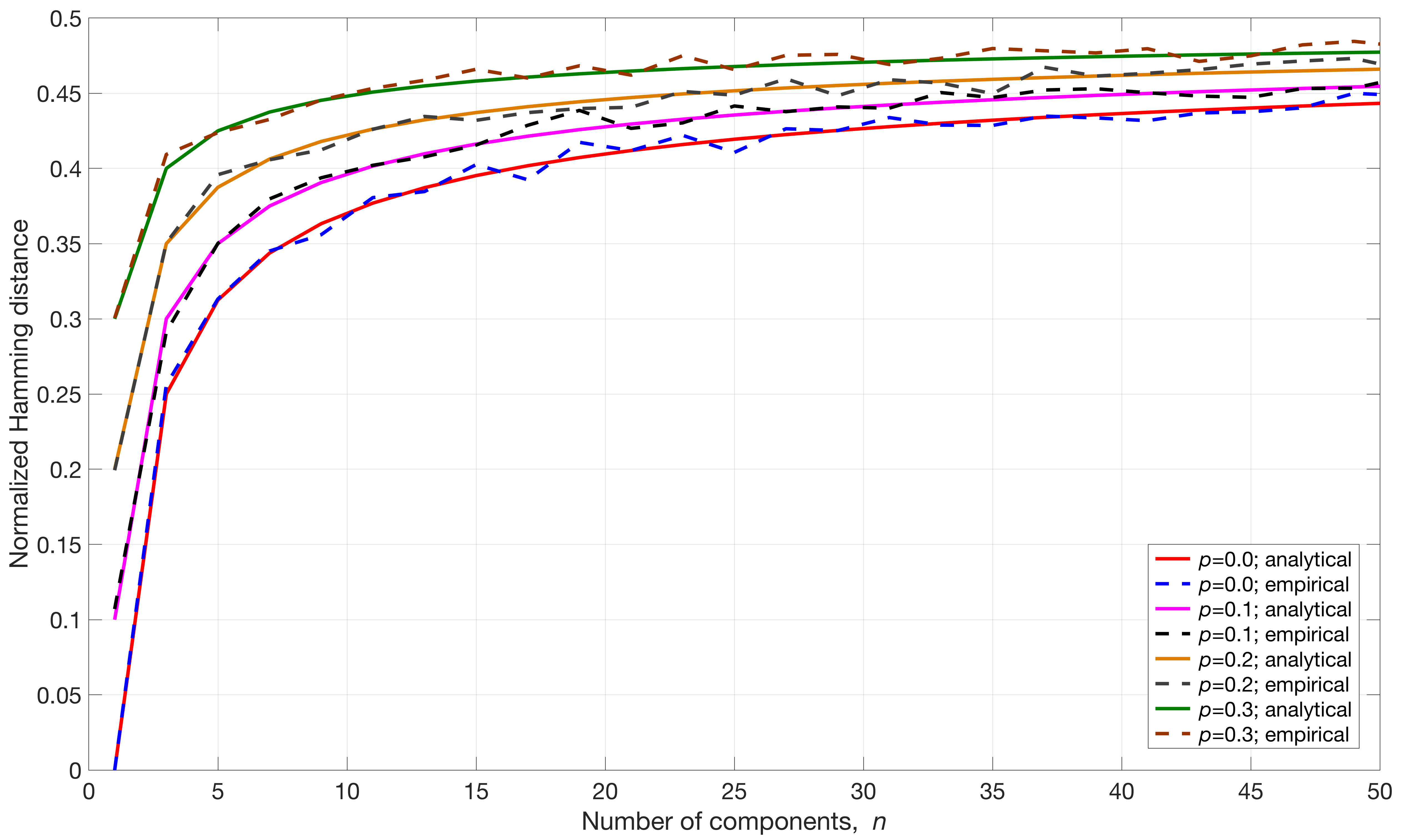}
\caption{
Normalized Hamming distance when probing a component HD vector from the result of the bundling operation, which is implemented via the consensus sum (majority rule). 
The figure shows both curves for analytical expression (\ref{eq:cap:correct:simplified}) as well as empirical curves for simulations. 
Four different values of $p$ were used: $[0.0, 0.1, 0.2, 0.3]$; $n$ varied in the range $[1,51]$.
} 
\label{fig:Hamming} 
\end{center}
\end{figure}

The original equation in section ``Theoretical limits on capacity of HBVs''~\cite{Mitrokhin19HD} for the expected normalized Hamming distance after the bundling operation is:
\noindent
\begin{equation}
(1-p) \sum_{k=n/2}^{n-1} \frac{(n-1)!}{ 2^{n-1}k!(n-k-1)!} + p\sum_{k=n/2-1}^{n-1} \frac{(n-1)!}{ 2^{n-1}k!(n-k-1)!},
\label{eq:cap:sr}
 \end{equation}
\noindent
where $n$ is the number of component HD vectors involved in the bundling operation and $p$ is the probability of a bit flip in either a component HD vector or in the result of the bundling operation.
However, when using the consensus sum (majority rule) $n$ should be odd (when $n$ is even it is implicit since ties are broken randomly). 
In this case, $k=n/2$ is  not an integer, therefore, indices in (\ref{eq:cap:sr}) should be written as:
\noindent
\begin{equation}
(1-p) \sum_{k=\frac{n-1}{2}+1}^{n-1} \frac{(n-1)!}{ 2^{n-1}k!(n-k-1)!} + p\sum_{k=\frac{n-1}{2}}^{n-1} \frac{(n-1)!}{ 2^{n-1}k!(n-k-1)!}.
\label{eq:cap:correct}
 \end{equation}
\noindent
Note that $\frac{(n-1)!}{k!(n-k-1)!}$ is a binomial coefficient and can be denoted as $ \binom{n-1}{k}$ then (\ref{eq:cap:correct}) can be rewritten as:
\noindent
\begin{equation}
(1-p) \frac{1}{2^{n-1}} \sum_{k=\frac{n-1}{2}+1}^{n-1} \binom{n-1}{k} + p \frac{1}{2^{n-1}} \sum_{k=\frac{n-1}{2}}^{n-1} \binom{n-1}{k}.
\label{eq:cap:correct:bin}
 \end{equation}
\noindent
Moreover, we know that: 
\noindent
\begin{equation}
\frac{1}{2^{n-1}} \sum_{k=0}^{n-1} \binom{n-1}{k} =1. 
\label{eq:cap:sum}
 \end{equation}
\noindent
Since $n$ is odd we can also write (\ref{eq:cap:sum}) as: 
\noindent
\begin{equation}
\frac{1}{2^{n-1}}  \sum_{k=0}^{\frac{n-1}{2}-1} \binom{n-1}{k} +  \frac{1}{2^{n-1}}  \binom{n-1}{\frac{n-1}{2}} +  \frac{1}{2^{n-1}}  \sum_{k=\frac{n-1}{2}+1}^{n-1} \binom{n-1}{k} =1. 
\label{eq:cap:sum:many}
\end{equation}

Because binomial coefficients are symmetric (e.g., $\binom{n-1}{0}=\binom{n-1}{n-1}=1$) we can state that $\sum_{k=0}^{\frac{n-1}{2}-1} \binom{n-1}{k} = \sum_{k=\frac{n-1}{2}+1}^{n-1} \binom{n-1}{k}$. This allows us to modify (\ref{eq:cap:sum:many}) to:
\noindent
\begin{equation}
\frac{2}{2^{n-1}}  \sum_{k=\frac{n-1}{2}+1}^{n-1} \binom{n-1}{k} +  \frac{1}{2^{n-1}}  \binom{n-1}{\frac{n-1}{2}}  =1. 
\label{eq:cap:sum:two}
\end{equation}
\noindent
From~(\ref{eq:cap:sum:two}) we get the an alternative expression of the left term in~(\ref{eq:cap:correct:bin}):  
\noindent
\begin{equation}
\frac{1}{2^{n-1}}  \sum_{k=\frac{n-1}{2}+1}^{n-1} \binom{n-1}{k} = \frac{1}{2} -  \frac{1}{2^{n}}  \binom{n-1}{\frac{n-1}{2}}.
\label{eq:cap:left}
\end{equation}
\noindent
Note that when $p=0$, (\ref{eq:cap:correct:bin}) is equivalent to (\ref{eq:cap:left}).
This corresponds to the case when there is no external noise (i.e., bit flips) being present in HD vectors. 
This case was considered in~\cite{Kanerva97} where the right hand side of (\ref{eq:cap:left}) was presented.

Moreover, due to the symmetry of binomial coefficients we know that 
$\sum_{k=\frac{n-1}{2}}^{n-1} \binom{n-1}{k} = \sum_{k=0}^{\frac{n-1}{2}} \binom{n-1}{k}$, therefore, when combining this with (\ref{eq:cap:sum}) we can express the right term in~(\ref{eq:cap:correct:bin}) through the left term as:
\begin{equation}
\frac{1}{2^{n-1}} \sum_{k=\frac{n-1}{2}}^{n-1} \binom{n-1}{k}= 1- \frac{1}{2^{n-1}}  \sum_{k=\frac{n-1}{2}+1}^{n-1} \binom{n-1}{k} = \frac{1}{2} +  \frac{1}{2^{n}}  \binom{n-1}{\frac{n-1}{2}}.
\label{eq:cap:right}
\end{equation}
\noindent
Thus, using the simplified expressions from~(\ref{eq:cap:left}) and~(\ref{eq:cap:right}), (\ref{eq:cap:correct:bin}) can be simplified to: 
\noindent
\begin{equation}
(1-p) \left( \frac{1}{2} -  \frac{1}{2^{n}}  \binom{n-1}{\frac{n-1}{2}} \right) + 
 p \left( \frac{1}{2} +  \frac{1}{2^{n}}  \binom{n-1}{\frac{n-1}{2}} \right),
\label{eq:cap:correct:simple}
 \end{equation}
\noindent
which eventually can be written  as: 
\noindent
\begin{equation}
\frac{1}{2} - \frac{1-2p}{2^{n}} \binom{n-1}{\frac{n-1}{2}}.
\label{eq:cap:correct:simplified}
 \end{equation}
\noindent
Note that~(\ref{eq:cap:correct:simplified}) is also valid in the case (shown in equation (5) in~\cite{HDNP17}) when $p$ characterizes bit flip probability in the resultant HD vector (after bundling) while components in the item memory are noise-free.
For visualization purposes, Fig.~\ref{fig:Hamming} presents the analytical solution in~(\ref{eq:cap:correct:simplified}) and results of simulations for several different values of $p$.

Importantly,~(\ref{eq:cap:correct:simplified}) is not sufficient to calculate the information capacity of HD vectors as it only provides the expected normalized Hamming distance.
The capacity will also depend on such parameters as the dimensionality of HD vectors and the size of item memory, which stores component (atomic) HD vectors. 
Early results on the capacity were given in~\cite{PlateThesis, PlateBook}. 
Some ideas for the case of binary/bipolar HD vectors were also presented in~\cite{Kleyko2015, Gallant13}.
Arguably the most comprehensive analysis of the capacity of different VSAs frameworks and even some classes of recurrent artificial neural networks has been recently presented in~\cite{Frady17}. 
Additionally,~\cite{Summers2018} elaborates on methods of recovering information from composite HD vectors.
Last, it is worth noting that the result of the bundling operation should not necessarily be used as system's memory. 
The resultant HD vectors could, in turn, be stored in some associative memory.
For example, Sparse Distributed Memory~\cite{KanervaBook} is a natural candidate for that.

\IEEEpeerreviewmaketitle

\bibliographystyle{ieeetr}
\bibliography{Bibliography}

\end{document}